\documentclass[10pt,twocolumn,letterpaper]{article}

\usepackage{cvpr}              

\usepackage[accsupp]{axessibility} 
\usepackage{amssymb}
\usepackage{pifont}
\usepackage{makecell}
\usepackage{graphicx}
\usepackage{everypage}   
\usepackage[utf8]{inputenc}  
\DeclareUnicodeCharacter{2217}{*} 
\newcommand{\xmark}{\ding{55}}%
\usepackage{multirow}
\definecolor{cvprblue}{rgb}{0.21,0.49,0.74}
\usepackage[pagebackref,breaklinks,colorlinks,allcolors=cvprblue]{hyperref}

\def\paperID{7502} 
\def\confName{CVPR}
\def\confYear{2025}

\title{Learning from Synchronization: Self-Supervised Uncalibrated Multi-View Person Association in Challenging Scenes}

\author{Keqi Chen$^{1}$ \quad  Vinkle Srivastav$^{1,2}$ \quad Didier Mutter$^{2,3}$ \quad  Nicolas Padoy$^{1,2}$  \vspace{0.3em} \\
{\normalsize $^1$University of Strasbourg, CNRS, INSERM, ICube, UMR7357, Strasbourg, France} \\
{\normalsize $^2$IHU Strasbourg, France} \\
{\normalsize $^3$University Hospital of Strasbourg, France} \\
{\tt\small \{keqi.chen, srivastav, npadoy\}@unistra.fr \quad didier.mutter@chru-strasbourg.fr}
}

\AddEverypageHook{%
    \ifnum\value{page}=1
        \rlap{\raisebox{6mm}{\hspace*{6mm}\small This article has been accepted to the IEEE/CVF Conference on Computer Vision and Pattern Recognition (CVPR), 2025.}}%
    \fi
}

\begin{document}
\maketitle
\begin{abstract}
Multi-view person association is a fundamental step towards multi-view analysis of human activities. Although the person re-identification features have been proven effective, they become unreliable in challenging scenes where persons share similar appearances. Therefore, cross-view geometric constraints are required for a more robust association. However, most existing approaches are either fully-supervised using ground-truth identity labels or require calibrated camera parameters that are hard to obtain. In this work, we investigate the potential of learning from synchronization, and propose a self-supervised uncalibrated multi-view person association approach, \emph{Self-MVA}, without using any annotations. Specifically, we propose a self-supervised learning framework, consisting of an encoder-decoder model and a self-supervised pretext task, cross-view image synchronization, which aims to distinguish whether two images from different views are captured at the same time. 
The model encodes each person's unified geometric and appearance features, and we train it by utilizing synchronization labels for supervision after applying Hungarian matching to bridge the gap between instance-wise and image-wise distances.
To further reduce the solution space, we propose two types of self-supervised linear constraints: multi-view re-projection and pairwise edge association. Extensive experiments on three challenging public benchmark datasets (WILDTRACK, MVOR, and SOLDIERS) show that our approach achieves state-of-the-art results, surpassing existing unsupervised and fully-supervised approaches. Code is available at \url{https://github.com/CAMMA-public/Self-MVA}.
\end{abstract}    
\section{Introduction}
\label{sec:intro}

\begin{figure*}
  \centering
  \includegraphics[width=1.0\linewidth]{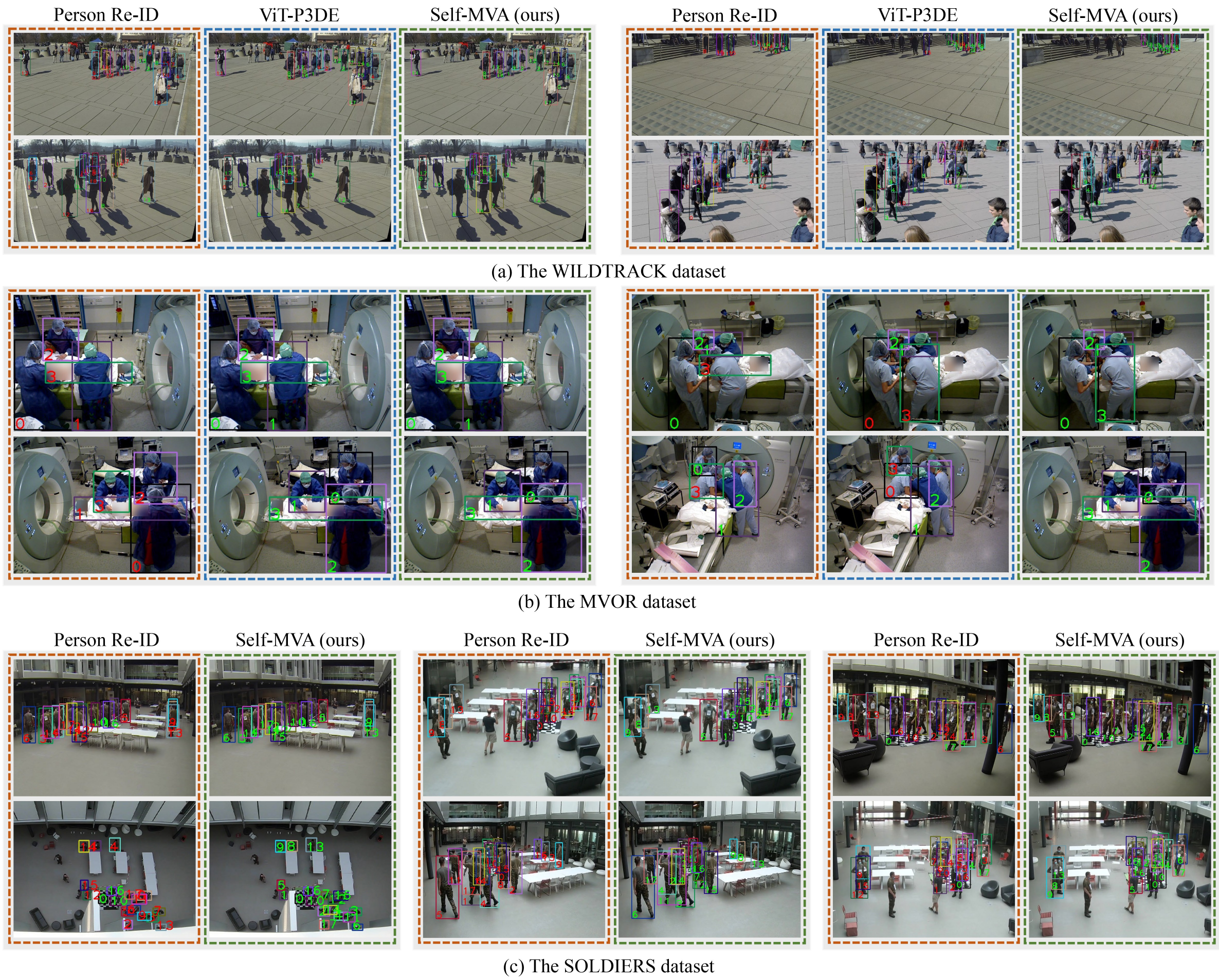}
  \caption{Examples of different multi-view person association methods on the challenging WILDTRACK \cite{chavdarova2018wildtrack}, MVOR \cite{srivastav2018mvor} and SOLDIERS \cite{soldiers} datasets. ViT-P3DE \cite{seo2023vit} is fully-supervised, while Person Re-ID \cite{zhou2021learning} and Self-MVA are unsupervised. Red and green identities mean incorrect and correct association respectively. Best viewed in color.}
  \label{fig:intro_compare}
\end{figure*}

The task of multi-view person association aims to associate the same persons in different views given the synchronized multi-view RGB images with overlapping fields of view. It is a fundamental step for many downstream computer vision tasks, including multi-view detection~\cite{chavdarova2018wildtrack,song2021stacked}, tracking~\cite{cheng2023rest,yang2024unified}, and 3d pose estimation~\cite{dong2019fast,chen2020multi}. To tackle this problem, previous works mostly follow the idea of traditional person Re-IDentification (Re-ID) by training or using an off-the-shelf person Re-ID model for affinity measurement~\cite{dong2019fast,xu2022multi}. Some works also utilize the unique properties of multi-view data by introducing multi-view cycle consistency and cross-view epipolar constraint~\cite{dong2019fast,cai2020messytable} as complementary cues, which make the association more robust.

Despite the remarkable progress of existing multi-view association research, current approaches do not perform well in challenging scenes where persons share similar appearances.~\cref{fig:intro_compare} shows examples of some challenging benchmark datasets: (1) in the WILDTRACK dataset~\cite{chavdarova2018wildtrack}, the diverse illumination in the wild, the crowdedness and the dramatic view change make the cross-view association extremely difficult even for the human eye; (2) in the MVOR~\cite{srivastav2018mvor} and SOLDIERS datasets~\cite{soldiers}, the clinicians and the soldiers all wear similar outfits, resulting in lack of individual characteristics. To visualize the challenges, we show the examples of applying existing association approaches on those datasets, including (1) a well-trained person Re-ID model OSNet~\cite{zhou2021learning} with Hungarian matching~\cite{kuhn1955hungarian}; (2) the fully-supervised instance association approach ViT-P3DE~\cite{seo2023vit}, encoding both context and geometric information through Vision Transformer~\cite{dosovitskiy2020vit}. As shown in~\cref{fig:intro_compare}, the person Re-ID model fails to distinguish similarly-looking persons, and ViT-P3DE makes wrong predictions of persons who stand too close. 

Furthermore, current research mostly concentrates on fully-supervised approaches, which require labor-intensive bounding box and identity annotations, and do not generalize well to unseen datasets. A few works have explored self-supervised association through cycle consistency~\cite{wang2020cycas,gan2021self}, but are limited to training a visual descriptor without explicitly incorporating geometry into account, and thus fail to address challenging scenes as illustrated in~\cref{fig:intro_compare}. Also, several works introduce cross-view geometric constraints to reduce the solution space~\cite{cai2020messytable,seo2023vit,dong2019fast}, but most of them require calibrated camera parameters, which are hard to obtain in practical scenarios. 

\begin{figure}[t]
  \centering
   \includegraphics[width=1.0\linewidth]{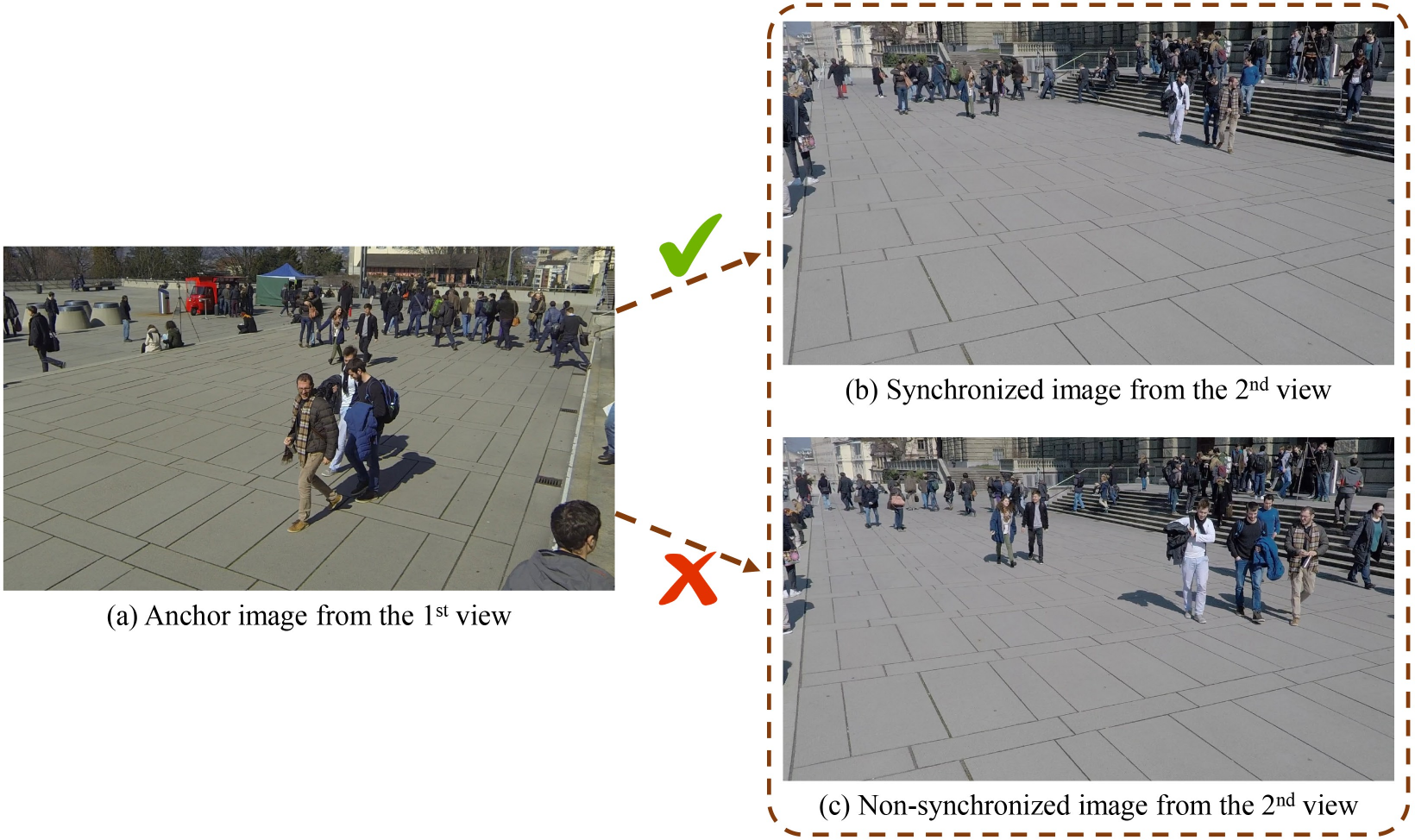}
   \caption{Examples of synchronized and non-synchronized image pairs in the WILDTRACK dataset \cite{chavdarova2018wildtrack}.}
   \label{fig:intro}
\end{figure}

Inspired by the idea of learning from multi-view synchronization~\cite{yilmaz2006object,jenni2020self}, we propose to learn cross-view geometric correspondences in a self-supervised way by determining whether two images of different views are captured at the same time. As shown in~\cref{fig:intro}, the synchronized and non-synchronized images in the second view manifest slightly different distributions of the persons' spatial locations. By explicitly modeling the distributions and distinguishing such tiny deviations of the scene layouts, the model can learn to encode fine-grained geometric representations of each person in a common feature space. 

We propose a novel self-supervised approach for multi-view person association, \textit{Self-MVA}, that exploits multi-view synchronization to learn geometrically constrained person associations across views in challenging scenes without requiring any annotated data. Specifically, we first propose an encoder-decoder model to obtain instance representations. The encoder maps each person's geometric information onto a common feature space for unified representation, and the decoder projects the geometric features back to each view to predict the 2d positions in the original view. Then, to train the model, we propose a self-supervised pretext task named cross-view image synchronization, which aims to distinguish whether two images from different views are captured at the same time. Specifically, we consider an image in the first view as the anchor image, and then use the corresponding synchronized image and a randomly selected non-synchronized image in the second view as the positive and negative samples respectively, as shown in~\cref{fig:intro}. With the obtained labels of image pairs and the embeddings of each person, we use Hungarian matching to conduct cross-view person association and compute the minimum averaged instance-wise distances as the image-wise distance, followed by using the triplet loss for training. 

Although cross-view image synchronization is theoretically capable of learning instance representation using only image-level supervision, it is too challenging for the model to converge in crowded scenes due to the enormous solution space. Therefore, in order to reduce the solution space, we propose two types of self-supervised linear constraints: \emph{the multi-view re-projection} and \emph{the pairwise edge association}. 

The multi-view re-projection is motivated by cycle consistency~\cite{zhu2017unpaired}, which aims to predict the persons' 2d locations in each view by re-projection given the geometric embeddings. Specifically, we use decoders of linear layers for each view to conduct linear regression, obtaining the estimated bounding boxes. Then, we use the detected bounding boxes in each view as the pseudo labels for instance-level supervision, which can prevent potential information loss during feature encoding. 

The pairwise edge association is an extension of the person association, which is inspired by~\cite{chen2020monopair}. The core idea is that considering multi-view images as multiple graphs where each node represents a person, the edges in different views should also be associated if the adjacent nodes are associated correspondingly. Therefore, we create a pseudo person who stands at two persons' geometric center as the pseudo ``edge'', and then we compute the edge-based image-wise distance for cross-view image synchronization. The pairwise edge association explicitly imposes linear constraints on the geometric representation, which helps the model achieve better performance. 

We evaluate our self-supervised approach on three challenging benchmark datasets, WILDTRACK \cite{chavdarova2018wildtrack}, MVOR \cite{srivastav2018mvor} and SOLDIERS \cite{soldiers}, and conduct extensive ablation studies. Experimental results show that our approach outperforms both unsupervised and fully-supervised approaches, proving its effectiveness. We also provide a qualitative analysis of the potential applications of our approach. 

We summarize our contributions as follows: (1) We address the multi-view person association in the challenging scenes using a self-supervised approach, without using any annotations, which surpasses the state-of-the-art unsupervised and fully-supervised approaches. (2) We introduce a self-supervised learning framework, including an encoder-decoder model and a self-supervised pretext task named cross-view image synchronization. (3) We propose two types of self-supervised linear constraints to reduce the solution space and optimize the model training, including the multi-view re-projection and the pairwise edge association.


\section{Related work}
\label{sec:related}

\subsection{Multi-view person association}

The traditional approach solves multi-view person association by minimizing the geometric error between the estimated 3d poses and the input 2d poses among all hypotheses~\cite{hartley2003multiple,belagiannis20143d,belagiannis20153d,ershadi2018multiple,dong2021shape}. It is computationally expensive and highly sensitive to noisy 2d poses. Apart from 2D-to-3D human pose lifting, recent works tend to solve the association at the bounding-box level with looser constraints in a fast and robust way~\cite{dong2019fast}. They usually conduct deep learning-based feature encoding along with affinity measurement for the association. For instance, they usually train a visual descriptor~\cite{vo2020self,gan2021self} or utilize an off-the-shelf person Re-ID model~\cite{dong2019fast,luna2022graph} to encode a person's appearance features. For multi-view videos, some works use an object tracker to obtain tracklets in each view for tracklet-based association~\cite{vo2020self}, while some works perform motion-level association through optical flow and skeleton features~\cite{zheng2017learning}. Considering the importance of geometric constraints across views, recent works propose to encode the geometric information for each person~\cite{han2021multiple,han2022connecting,seo2023vit}. In this work, we also encode fine-grained geometric features for detection-based person association in multi-view images. 

After feature encoding, the association can be divided into supervised and unsupervised approaches by whether they use ground-truth identity labels or not. The supervised approaches either conduct metric learning through contrastive and triplet losses \cite{zheng2017learning,seo2023vit}, or 
predict the assignment matrix through recurrent neural networks or graph neural networks~\cite{ardeshir2016ego2top,luna2022graph,zhang20204d,han2022multi}. The unsupervised approaches can be classified into clustering-based and linear assignment-based approaches. Clustering-based approaches solve this problem as constrained optimization, with the knowledge of the number of persons in the scene and view-specific constraints \cite{xu2022multi,xu2023multi,zhou2022quickpose}. Linear assignment-based approaches usually formulate the association as a maximum bipartite matching problem, and solve it using the corresponding algorithms \cite{gan2021self,cai2020messytable,chen2020multi}. Our approach is also based on linear assignment, which does not require annotations or scene-related knowledge. 

Given calibrated camera parameters, different approaches utilize geometric constraints for a more robust association. Some works utilize epipolar constraints to filter out unreasonable associations~\cite{cai2020messytable} or to calculate ray-based geometric affinity~\cite{chen2020cross,bridgeman2019multi}. Some works use a homography matrix to map the persons' locations onto the world ground plane to measure real-world distance \cite{chen2020multi,hu2022multi,luna2022graph,cheng2023rest}. These approaches all require accurate camera poses that are hard to obtain. In contrast, our approach incorporates geometric constraints implicitly, without using camera parameters. 



\subsection{Self-supervised multi-view learning}

In order to remove the dependency on the ground-truth identity labels, a few recent works concentrate on self-supervised multi-view person association. Some work utilizes mutual exclusive constraints and multi-view geometry to automatically generate triplets for metric learning \cite{vo2020self}. MvMHAT \cite{gan2021self} proposes the symmetric-similarity and transitive-similarity losses to train the encoder based on cycle consistency. However, these works are still limited to training a visual descriptor, which cannot distinguish persons of similar appearance. 

Apart from person association, the most related self-supervised learning approach utilizes multi-view synchronization itself as labels to exploit view-invariant features \cite{jenni2020self}. For multiple views of the same person, they train the model to implicitly learn the possible geometric rigid transformation, by distinguishing whether these two images of different views are captured at the same time. In this work, we extend their work to the scene-level domain: by distinguishing whether two views of the same scene are captured at the same time, our model learns the fine-grained instance-level geometric representation across views. 


\section{Methodology}
\label{sec:methodology}

\begin{figure*}
  \centering
  \includegraphics[width=0.9\linewidth]{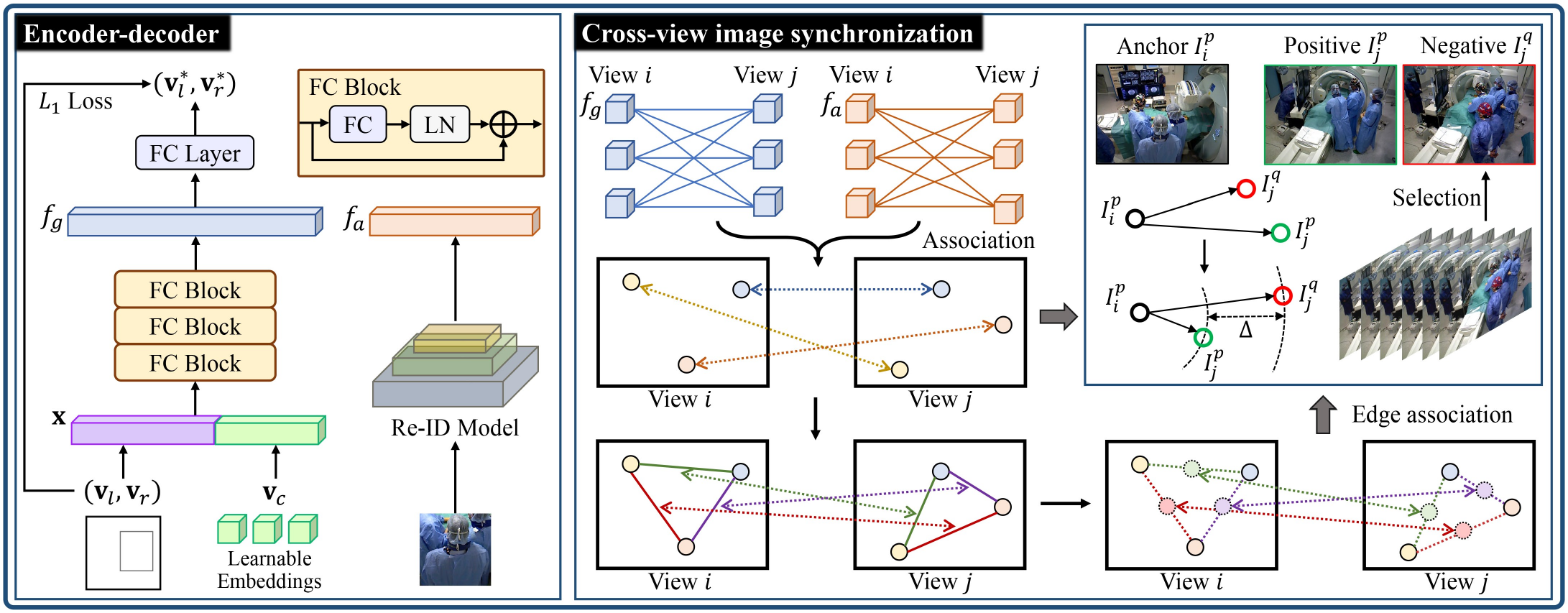}
  \caption{The framework of the self-supervised learning framework (FC = Fully-Connected, LN = Layer Normalization). For each image with detected persons, we encode each person's appearance features using a person Re-ID model, map their geometric information to a unified geometric feature space using positional encodings and learnable camera embeddings, and then decode the original 2d position. Then, we construct a triplet for each anchor image by randomly selecting a negative sample and conducting metric learning after instance association and edge association. Best viewed in color. }
  \label{fig:method}
\end{figure*}

\subsection{Problem overview}

Given a training dataset of multi-view human-centric RGB images $\mathcal{D} = \left\{ \mathbf{x}| \mathbf{y^{\ast}} \right\}$ where $ \mathbf{x} \in \mathcal{R}^{C \times 3 \times H \times W}$ is a multi-view image set from $C$ cameras with height $H$ and width $W$, and $\mathbf{y^{\ast}} \in \mathcal{R}^{C \times P \times 4}$ represents the 2d pseudo bounding boxes for $P$ persons generated by an off-the-shelf detector~\cite{zheng2022progressive}, the goal is to train a deep learning model that can encode distinguishable features of each person $\mathcal{Y} \in \mathcal{R}^{C \times P \times E}$ for detection-based multi-view association, where $E$ represents the embedding dimension. It should be noted that $P$ can vary in each camera view due to occlusion and noisy detections. For simplicity in the notation, we keep the same variable $P$. 

In the following, we describe our self-supervised learning framework. We first use an off-the-shelf detector~\cite{zheng2022progressive} to obtain the persons' bounding boxes for training. Then, we propose a multi-view learning framework as shown in \cref{fig:method}, which consists of an encoder-decoder model and a self-supervised pretext task named cross-view image synchronization. Furthermore, we propose two types of self-supervised linear constraints for better model training. We introduce each key step in detail as follows. 

\subsection{Encoder-decoder model}
\label{sec:encoder_decoder}

To obtain a unified representation of the persons in different views, we propose an encoder-decoder model, as shown in~\cref{fig:method}. The encoder encodes both the appearance features and the high-dimensional geometric features of the persons in different views. The decoder projects the geometric features back to each view to obtain the 2d positions, which is only used during training. 

With the obtained bounding boxes of all the persons in each view, we first represent the boxes by positional encodings~\cite{tancik2020fourier}. Specifically, we map the normalized top-left and bottom-right corner point positions $(\mathbf{v}_l, \mathbf{v}_r)$ to a higher dimensional hypersphere as follows, where $\mathbf{b}_j (1 \leq j \leq N)$ are the learnable Fourier basic frequencies:
\begin{equation}
\gamma(\mathbf{v}) = [\sin \mathbf{f}_1, \cos \mathbf{f}_1, ...,
\sin \mathbf{f}_N, \cos \mathbf{f}_N]^\mathrm{T},
 \label{eq:pos_encoding1}
\end{equation}
\begin{equation}
\mathbf{f}_j = 2\pi \mathbf{b}_j^{\mathrm{T}}\mathbf{v}.
 \label{eq:pos_encoding2}
\end{equation}

Then, to enable view-invariant geometric feature encoding without using camera parameters, we maintain learnable camera embeddings $\mathbf{v}_c \in \mathcal{R}^{C \times V}$ of size $V$ for $C$ views. For each person, we concatenate its 2d box representation and the camera embedding and pass it through fully-connected (FC) blocks $\mathrm\mathtt{F}(\cdot)$ to obtain the geometric features $f_{g}$:
\begin{equation}
\mathbf{x} = [\gamma(\mathbf{v}_l), \gamma(\mathbf{v}_r), \mathbf{v}_c],
 \label{eq:pos_in}
\end{equation}
\begin{equation}
f_{g} = \mathrm\mathtt{F}(\mathbf{x}).
 \label{eq:pos_out}
\end{equation}

Additionally, we also encode the appearance features $f_{a}$ through a person Re-ID model \cite{zhou2021learning}. In \cref{sec:syn}, we use it as a regularization term for feature association. 

Lastly, given the encoded geometric features $f_{g}$ of size $G$, we pass it through the decoder to localize the 2d positions in the original view by re-projection. Specifically, we use a single linear layer as the decoder for each view to obtain the estimated 2d positions of each box $(\mathbf{v}_l^{\ast}, \mathbf{v}_r^{\ast})$, where $W \in \mathcal{R}^{4 \times G}$ and $B \in \mathcal{R}^{4}$ are the weights and biases: 
\begin{equation}
(\mathbf{v}_l^{\ast}, \mathbf{v}_r^{\ast}) = Wf_{g} + B.
 \label{eq:decoder}
\end{equation}
In \cref{sec:constraints}, we use the decoder to apply linear constraints. 

\subsection{Cross-view image synchronization}
\label{sec:syn}

In order to train the geometric encoder in a self-supervised way, we propose to learn from multi-view synchronization. Motivated by~\cite{jenni2020self}, where they use the knowledge that the human pose in synchronized two views should follow a rigid transformation, we extend it to the scene-level domain: the position distribution of the persons in two views should also follow a rigid transformation. Therefore, we introduce a pretext task to distinguish whether two images in different views are captured at the same time, through explicit modeling of the persons' positions. 

For each image $I$ with $P$ persons, we encode their appearance features $\mathcal{F}_a=\left\{ f^1_{a}, ..., f^P_{a} \right\}$ and geometric features $\mathcal{F}_g=\left\{ f^1_{g}, ..., f^P_{g} \right\}$. Given the $p^{\mathrm{th}}$ frame from view $i$, $I^p_i$, with $P^p_i$ persons, and the $q^{\mathrm{th}}$ frame from view $j$, $I^q_j$, with $P^q_j$ persons, we compute the normalized appearance and geometric Euclidean distance matrices $(\mathcal{D}_a, \mathcal{D}_g)\in \mathcal{R}^{P^p_i \times P^q_j}$, and then take their weighted sum as the overall distance matrix $\mathcal{D}$, representing instance-wise distances:
\begin{equation}
\mathcal{D} = \alpha \mathcal{D}_a + (1 - \alpha) \mathcal{D}_g.
 \label{eq:dis}
\end{equation}

To bridge the gap between instance-wise and image-wise distances, we consider the instance association between two images as a maximum bipartite matching problem and solve it using Hungarian matching. We thus obtain the matched row and column indices $(\mathcal{M}_r, \mathcal{M}_c)\in \mathcal{R}^{m}$ of matrix $\mathcal{D}$, where $m=\min(P^p_i,P^q_j)$. Afterwards, we average the pairwise instance-wise distances as the image-wise distance $\mathcal{H}$: 
\begin{equation}
\mathcal{H} = \frac{1}{m}\sum \mathcal{D}[\mathcal{M}_r; \mathcal{M}_c].
 \label{eq:img_dis}
\end{equation}

Finally, we conduct triplet-based metric learning. Given an anchor image $I^p_i$, we consider $(I^p_i, I^p_j)$ as the positive pair, and $(I^p_i, I^q_j) (q\neq p)$ as the negative pair. Specifically, $q$ is randomly chosen around $p$ that satisfies $\left|q-p\right| \in [t_{\mathrm{min}},t_{\mathrm{max}}]$, where $[t_{\mathrm{min}},t_{\mathrm{max}}]$ is manually-defined frame range. Then we compute the triplet loss $L_{\mathrm{syn}}$ for training:
\begin{equation}
L_{\mathrm{syn}}=\max (0, \mathcal{H}_{I^p_i,I^p_j} - \mathcal{H}_{I^p_i,I^q_j} + \triangle),
 \label{eq:triplet_syn}
\end{equation}
where $\triangle$ is the margin between positive and negative pairs. 

During inference, we also use Hungarian matching to associate the persons. It is to be noted that treating the association as a maximum bipartite matching problem may introduce false positive matches when some persons only appear in one view. Therefore, we compute the confidence scores $\mathcal{S}$ of matrix $\mathcal{D}$, and filter out the matches with low scores:
\begin{equation}
\mathcal{S} = (1 - \frac{\mathcal{D}}{\max \mathcal{D}}).
 \label{eq:conf}
\end{equation}

\subsection{Self-supervised linear constraints}
\label{sec:constraints}

Although the cross-view image synchronization task imposes strong constraints on the position distribution of the persons in different views, it does not directly provide instance-level supervision, and thus the model struggles to find the correct gradient descent direction when there are too many possible matches in the crowded scenes. Therefore, in order to optimize the training, we introduce two additional self-supervised linear constraints: the multi-view re-projection and the pairwise edge association. 

The multi-view re-projection is inspired by cycle consistency \cite{zhu2017unpaired}. As mentioned in \cref{sec:encoder_decoder}, we map the detected bounding boxes $(\mathbf{v}_l, \mathbf{v}_r)$ to the high-dimensional feature space through the encoder, obtaining geometric features $f_{g}$. In order to prevent potential information loss during feature encoding, we project $f_{g}$ back to the original view, obtaining the estimated bounding boxes $(\mathbf{v}_l^{\ast}, \mathbf{v}_r^{\ast})$. Then, we use $(\mathbf{v}_l, \mathbf{v}_r)$ as the pseudo labels, and compute the $L_1$ loss as the re-projection loss $L_{\mathrm{pro}}$:
\begin{equation}
L_{\mathrm{pro}}=\mathcal{L}_1(\mathbf{v}_l^{\ast}, \mathbf{v}_l) + \mathcal{L}_1(\mathbf{v}_r^{\ast}, \mathbf{v}_r).
 \label{eq:loss_loc}
\end{equation}

Inspired by \cite{chen2020monopair}, we propose the pairwise edge association. As shown in \cref{fig:method}, we consider multi-view images as multiple graphs, where each node represents a person. \cref{sec:syn} has conducted the node association to compute the image-wise distance. Based on the node association, we further propose the pairwise edge association to measure the image-wise distance from another angle. Specifically, if node $n_i^u$ and node $n_i^v$ in the $i^{\mathrm{th}}$ view are associated with node $n_j^u$ and node $n_j^v$ in the $j^{\mathrm{th}}$ view respectively, then the edge $e_i^{u,v}$ in the $i^{\mathrm{th}}$ view and the edge $e_j^{u,v}$ in the $j^{\mathrm{th}}$ view should also be associated. However, we do not really have the physical edges, and thus for every two nodes $(\mathbf{v}_l^u, \mathbf{v}_r^v)$ and $(\mathbf{v}_l^u, \mathbf{v}_r^v)$ in the same view, we create a pseudo person $(\mathbf{v}_l^{u,v}, \mathbf{v}_r^{u,v})$ who stands at their geometric center as the pseudo ``edge'': 
\begin{equation}
(\mathbf{v}_l^{u,v}, \mathbf{v}_r^{u,v}) = (\frac{\mathbf{v}_l^u + \mathbf{v}_l^v}{2}, \frac{\mathbf{v}_r^u + \mathbf{v}_r^v}{2}).
 \label{eq:pseudo_edge}
\end{equation}
Then, we pass the pseudo ``edge'' through the encoder, obtaining its geometric features $f_g^{u,v}$. 

The above approximation is based on the knowledge that the positions of the persons in two views should follow a rigid transformation. In other words, the feature encoding transformation $\mathcal{E}(\cdot)$ preserves linear combinations:
\begin{equation}
f_{g}^{u,v} = \mathrm\mathtt{\mathcal{E}}(\mathbf{v}_l^{u,v}, \mathbf{v}_r^{u,v}) = \frac{1}{2}\mathrm\mathtt{\mathcal{E}}(\mathbf{v}_l^{u}, \mathbf{v}_r^{u}) + \frac{1}{2}\mathrm\mathtt{\mathcal{E}}(\mathbf{v}_l^{v}, \mathbf{v}_r^{v}).
 \label{eq:pos_out_edge}
\end{equation}

Suppose that we have $m$ node matches as computed in \cref{sec:syn}, we thus have $\hat{m} = m(m-1)/2$ edge matches. Then we compute the edge-based image-wise distance $\hat{\mathcal{H}}$ by averaging the Euclidean distances as follows:
\begin{equation}
\hat{\mathcal{H}}_{i,j} = \frac{1}{\hat{m}}\sum \mathcal{L}_2(f_{g,i}^{u,v}, f_{g,j}^{u,v}).
 \label{eq:img_dis_edge}
\end{equation}

Afterwards, similar to \cref{eq:triplet_syn}, we compute the edge-based triplet loss for each triplet $(I^p_i,I^p_j,I^q_j)$:
\begin{equation}
L_{\mathrm{edge}}=\max (0, \hat{\mathcal{H}}_{I^p_i,I^p_j} - \hat{\mathcal{H}}_{I^p_i,I^q_j} + \triangle).
 \label{eq:triplet_edge}
\end{equation}

In general, the final loss function is as follows: 
\begin{equation}
L=L_{\mathrm{syn}} + L_{\mathrm{pro}} + L_{\mathrm{edge}}.
 \label{eq:loss}
\end{equation}

\subsection{Implementation details}

\paragraph{\textbf{Training strategies}} We use the PyTorch framework to implement our approach with a single NVIDIA A40 GPU on the Ubuntu system. We use the Adam optimizer with an initial learning rate of 1e-4, which decreases to 1e-5 in the end. Specifically, for the WILDTRACK, MVOR, and SOLDIERS datasets, we train the model for 1200, 80, and 320 epochs, and decrease the learning rate at the 900th, 60th, and 240th epochs respectively. For positional encoding, we set $N$ to 128 in \cref{eq:pos_encoding1}. The sampling range $[t_{\mathrm{min}},t_{\mathrm{max}}]$ in cross-view image synchronization is set to $[5,20]$. The $\alpha$ in \cref{eq:dis} is set to 0.1. The $\triangle$ in \cref{eq:triplet_syn} is set to 1.0. 

\paragraph{\textbf{Inference pipeline}} During inference, we use the encoder to obtain each person's embeddings. Then, we use the Hungarian matching algorithm to get the association. Finally, we filter out matches with low confidence scores in \cref{eq:conf}. 
The thresholds are set as 0.4, 0.1, and 0 for the WILDTRACK, MVOR, and SOLDIERS datasets respectively. 
\section{Experiments}
\label{sec:experiments}

\begin{table*}[t]
\centering
\resizebox{1.3\columnwidth}{!}{
\begin{tabular}{c|cccccccc}
    \toprule
    Methods & AP $\uparrow$ & FPR-95 $\downarrow$ & P $\uparrow$ & R $\uparrow$ & \textbf{ACC} $\uparrow$ & \textbf{IPAA-100} $\uparrow$ & \textbf{IPAA-90} $\uparrow$ & \textbf{IPAA-80} $\uparrow$ \\
    \midrule
    OSNet \cite{zhou2021learning} & 16.80 & 92.09 & 28.27 & 29.68 & 37.56 & 0.0 & 0.48 & 3.21 \\
    OSNet + ESC \cite{cai2020messytable} & 59.52 & 15.32 & 78.11 & 79.07 & 82.13 & 26.43 & 39.40 & 66.67 \\
    MvMHAT \cite{gan2021self} & 4.44 & 94.13 & 5.96 & 6.29 & 22.37 & 0.0 & 0.48 & 1.55 \\
    ASNet$^\ast$ \cite{cai2020messytable} & \textbf{73.40} & 8.30 & - & - & - & 32.10 & - & - \\
    GNN-CCA$^\ast$ \cite{luna2022graph} & 4.12 & 93.30 & - & 0.0 & 36.82 & 0.0 & 2.14 & 14.40 \\
    ViT-P3DE$^\ast$ \cite{seo2023vit} & 70.46 & \textbf{5.83} & 86.92 & 87.01 & 89.49 & 35.48 & 53.10 & 84.17 \\
    \midrule
    Self-MVA (ours) & 56.68 & 11.19 & \textbf{92.31} & \textbf{94.34} & \textbf{91.73} & \textbf{54.64} & \textbf{65.60} & \textbf{86.55} \\
    \bottomrule
\end{tabular}
}
\caption{Results on the WILDTRACK dataset (ESC = Epipolar Soft Constraint~\cite{cai2020messytable}). Methods with $^\ast$ are fully-supervised approaches.}
\label{tab:wildtrack}
\end{table*}

\begin{table}
\centering
\resizebox{0.9\columnwidth}{!}{
\begin{tabular}{c|cccccc}
    \toprule
    Methods & AP $\uparrow$ & FPR-95 $\downarrow$ & P $\uparrow$ & R $\uparrow$ & \textbf{ACC} $\uparrow$ & \textbf{IPAA-100} $\uparrow$ \\
    \midrule
    OSNet \cite{zhou2021learning} & 53.58 & 85.98 & 71.11 & 75.99 & 69.05 & 63.82 \\
    OSNet + ESC \cite{cai2020messytable} & 73.56 & 60.05 & 79.75 & 84.17 & 77.84 & 74.37 \\
    MvMHAT \cite{gan2021self} & 33.28 & 97.31 & 56.54 & 60.42 & 55.13 & 51.76  \\
    GNN-CCA$^\ast$ \cite{luna2022graph} & 84.70 & \textbf{44.45} & 92.13 & 52.51 & 65.38 & 52.76  \\
    ViT-P3DE$^\ast$ \cite{seo2023vit} & 81.32 & 78.97 & 92.45 & 93.67 & 89.19 & \textbf{84.92} \\
    \midrule
    Self-MVA (ours) & \textbf{86.50} & 79.44 & \textbf{93.20} & \textbf{93.93} & \textbf{89.38} & 83.92 \\
    \bottomrule
\end{tabular}
}
\caption{Results on the MVOR dataset (ESC = Epipolar Soft Constraint~\cite{cai2020messytable}). Methods with $^\ast$ are fully-supervised approaches.}
\label{tab:mvor}
\end{table}

\begin{table}
\centering
\resizebox{0.9\columnwidth}{!}{
\begin{tabular}{c|cccccc}
    \toprule
    Methods & AP $\uparrow$ & FPR-95 $\downarrow$ & P $\uparrow$ & R $\uparrow$ & \textbf{ACC} $\uparrow$ & \textbf{IPAA-100} $\uparrow$ \\
    \midrule
    OSNet \cite{zhou2021learning} & 14.97 & 95.13 & 23.75 & 23.75 & 23.75 & 4.61 \\
    OSNet + ESC \cite{cai2020messytable} & 56.90 & \textbf{29.53} & 79.91 & 79.91 & 79.91 & 52.12 \\
    MvMHAT \cite{gan2021self} & 13.52 & 94.75 & 22.06 & 22.06 & 22.06 & 3.15  \\
    \midrule
    Self-MVA (ours) & \textbf{79.13} & 32.26 & \textbf{95.89} & \textbf{95.89} & \textbf{95.89} & \textbf{87.64} \\
    \bottomrule
\end{tabular}
}
\caption{Results on the SOLDIERS dataset (ESC = Epipolar Soft Constraint~\cite{cai2020messytable}).}
\label{tab:soldiers}
\end{table}

\begin{table*}[t]
\centering
\resizebox{1.3\columnwidth}{!}{
\begin{tabular}{c|c|c|cccccccc}
    \toprule
    Syn. & Pro. & Edge. & AP $\uparrow$ & FPR-95 $\downarrow$ & P $\uparrow$ & R $\uparrow$ & \textbf{ACC} $\uparrow$ & \textbf{IPAA-100} $\uparrow$ & \textbf{IPAA-90} $\uparrow$ & \textbf{IPAA-80} $\uparrow$ \\
    \midrule
    \checkmark & \xmark & \xmark & 4.23 & 94.62 & 5.74 & 6.06 & 21.00 & 0.0 & 0.24 & 0.48 \\
    \xmark & \checkmark & \xmark & 4.40 & 95.60 & 6.03 & 6.37 & 22.83 & 0.0 & 0.0 & 1.07 \\
    \checkmark & \checkmark & \xmark & 54.67 & 14.48 & 86.92 & 87.75 & 86.15 & 24.05 & 41.31 & 73.45 \\
    \checkmark & \checkmark & \checkmark & \textbf{56.68} & \textbf{11.19} & \textbf{92.31} & \textbf{94.34} & \textbf{91.73} & \textbf{54.64} & \textbf{65.60} & \textbf{86.55} \\
    \bottomrule
\end{tabular}
}
\caption{Ablation study of the self-supervised learning tasks. Cross-view image synchronization (Syn.), multi-view re-projection (Pro.) and pairwise edge association (Edge.) are all necessary for the training.}
\label{tab:syn_loc_edge}
\end{table*}

\subsection{Datasets and evaluation metrics}

We conduct experiments on three benchmark datasets: WILDTRACK~\cite{chavdarova2018wildtrack}, MVOR~\cite{srivastav2018mvor} and SOLDIERS~\cite{soldiers}. For WILDTRACK and MVOR datasets, following previous works' standard \cite{luna2022graph,cheng2023rest}, we use 70\% data for training, 20\% for validation, and 10\% for testing. For SOLDIERS dataset, we use 10\% for testing and the rest for training. 

The WILDTRACK dataset is a multi-view video dataset with seven cameras capturing an unscripted dense group of pedestrians in a public open area. We use it as the main dataset for both comparison and ablation studies. 

The MVOR dataset is a multi-view image dataset in the operating room, with three cameras capturing clinicians during the surgical procedure. The original dataset does not have complete identity labels for every person, and thus we annotate all the identity labels ourselves for evaluation and the training of the fully-supervised methods. 

The SOLDIERS dataset is a multi-view video dataset with six cameras capturing soldiers walking around in the atrium of a building. 
The original annotations only use a single point to represent each person. Therefore, we annotate both bounding boxes and the identity labels of the test set ourselves for evaluation only. We did not annotate the full dataset due to the challenges of associating the soldiers across the views even for the human eye. 

We evaluate our approach with the following evaluation metrics: average precision (AP), false positive rate at 95\% recall (FPR-95), precision (P), recall (R), accuracy (ACC), and image pair association accuracy (IPAA-X). Precision and recall only compute the positive matches, while accuracy computes the overall association accuracy for both positive and negative matches. IPAA-X is computed as the fraction of image pairs that reach at least X\% accuracy, which is a more stringent indicator than AP \cite{cai2020messytable}.

\subsection{Comparison with the state-of-the-art}

We compare our method with both the fully-supervised and unsupervised approaches, including: (1) off-the-shelf OSNet \cite{zhou2021learning} with Hungarian matching, representing the performance of a well-trained person Re-ID model; (2) OSNet with optimal epipolar soft constraint~\cite{cai2020messytable}, using camera parameters; (3) self-supervised MvMHAT \cite{gan2021self}; (4) fully-supervised graph-based GNN-CCA \cite{luna2022graph,cheng2023rest}, which utilizes camera parameters; (5) fully-supervised ASNet, which is pre-trained on MessyTable dataset for better performance \cite{cai2020messytable}; (6) fully-supervised ViT-P3DE \cite{seo2023vit}, which simultaneously encodes appearance and geometric information. 

\cref{tab:wildtrack} shows the results on the WILDTRACK dataset. Our self-supervised method achieves the best performance, proving its effectiveness. It surpasses the fully-supervised ViT-P3DE on most indicators, especially the IPAA-100. It also demonstrates that AP and FPR-95 do not fully represent the model's capability on the association task because their computation does not consider the constraint that each person can only be associated one time in another view. 

\cref{tab:mvor} and \cref{tab:soldiers} show the results on the MVOR and SOLDIERS datasets. Since we only annotate the test set of the SOLDIERS dataset, we compare Self-MVA with the unsupervised approaches. On both datasets, Self-MVA achieves the best performance, proving its usage on indoor challenging scenes. It is also worth mentioning that as a graph-based edge prediction method that is structurally suitable for association, GNN-CCA has a much better FPR-95 on the MVOR dataset. However, it does not perform well on the other indicators, especially the recall rate. 

\subsection{Ablation study}

\paragraph{\textbf{Analysis of cross-view image synchronization and self-supervised linear constraints}} 
\label{abl_syn_pre}
We examine the importance of the tasks within the self-supervised learning framework on the WILDTRACK dataset, including the cross-view image synchronization, the multi-view re-projection, and the pairwise edge association. As shown in \cref{tab:syn_loc_edge}, the model fails to converge in crowded scenes without either synchronization or re-projection task, and the additional edge association task guarantees the best performance.

\begin{table}[t]
\centering
\resizebox{1.0\columnwidth}{!}{
\begin{tabular}{c|cccccccccc}
    \toprule
    Re-ID & AP $\uparrow$ & FPR-95 $\downarrow$ & P $\uparrow$ & R $\uparrow$ & \textbf{ACC} $\uparrow$ & \textbf{IPAA-100} $\uparrow$ & \textbf{IPAA-90} $\uparrow$ & \textbf{IPAA-80} $\uparrow$ \\
    \midrule
    \xmark & 56.36 & 11.14 & 90.50 & 93.13 & 90.50 & 46.79 & 62.38 & 83.81 \\
    Test & \textbf{57.93} & \textbf{11.10} & 91.24 & 93.64 & 91.02 & 47.02 & 65.24 & 84.76 \\
    Train \& Test & 56.68 & 11.19 & \textbf{92.31} & \textbf{94.34} & \textbf{91.73} & \textbf{54.64} & \textbf{65.60} & \textbf{86.55} \\
    \bottomrule
\end{tabular}
}
\caption{Ablation study of the Re-ID feature. We conduct experiments on three settings: (1) no usage; (2) only use it during testing; (3) use it in both training and testing stages.}
\label{tab:reid}
\end{table}

\begin{table}[t]
\centering
\resizebox{1.0\columnwidth}{!}{
\begin{tabular}{c|cccccccc}
    \toprule
    Methods & AP $\uparrow$ & FPR-95 $\downarrow$ & P $\uparrow$ & R $\uparrow$ & \textbf{ACC} $\uparrow$ & \textbf{IPAA-100} $\uparrow$ & \textbf{IPAA-90} $\uparrow$ & \textbf{IPAA-80} $\uparrow$ \\
    \midrule
    Self-MVA & 56.68 & 11.19 & 92.31 & 94.34 & 91.73 & 54.64 & 65.60 & 86.55 \\
    + box & 80.03 & 5.66 & \textbf{99.07} & 98.87 & \textbf{98.82} & 86.67 & \textbf{96.90} & \textbf{99.52} \\
    + box \& id & \textbf{89.77} & \textbf{1.23} & 98.56 & \textbf{99.03} & 98.60 & \textbf{87.38} & 96.43 & 99.40 \\
    \bottomrule
\end{tabular}
}
\caption{Ablation study of training our model using ground-truth bounding boxes (box) and identity labels (id).}
\label{tab:gt}
\end{table}

\begin{figure*}[t]
\centering
\includegraphics[width=1.8\columnwidth]{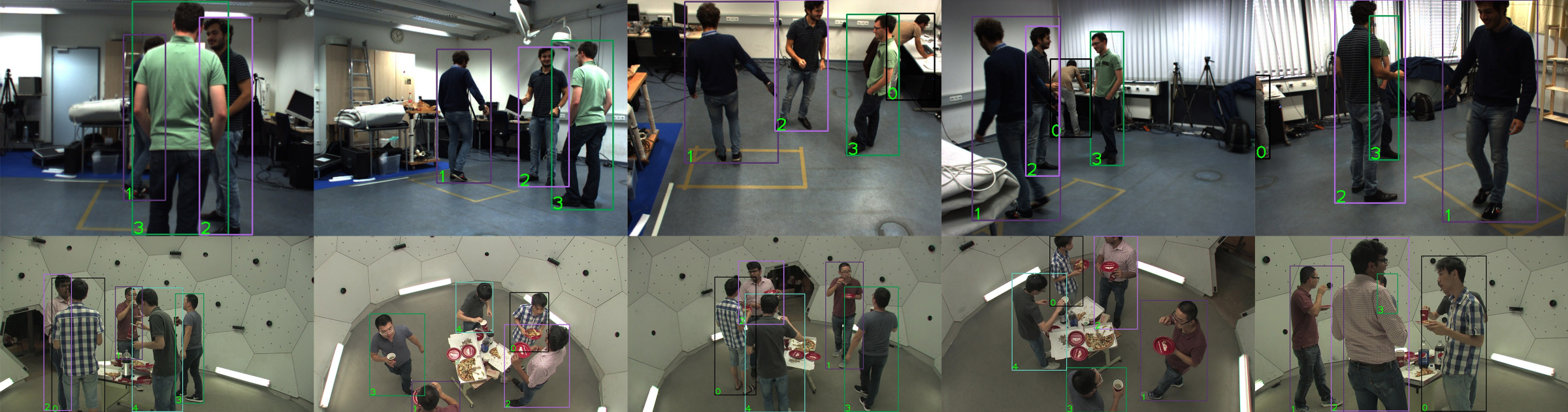} 
\caption{Example of multi-view person association on the Shelf \cite{belagiannis20143d} and Panoptic datasets \cite{joo2015panoptic}.}
\label{fig:pose}
\end{figure*}

\paragraph{\textbf{Analysis of the Re-ID feature}} 
\label{sec:hungarian_exp}
To evaluate the effectiveness of the Re-ID feature in the framework, we conduct an ablation study with three experiment settings: (1) no Re-ID feature; (2) only use the Re-ID feature in the testing stage; (3) use the Re-ID feature in both training and testing stage. As shown in \cref{tab:reid}, adding the Re-ID feature always improves the performance, and training with the Re-ID feature achieves the best performance. It is also worth mentioning that the model already achieves great performance without using the Re-ID feature, which further proves the significance of geometric features in challenging scenes.

\paragraph{\textbf{Training with ground-truth bounding boxes and identity labels}} The compared fully-supervised \cite{cai2020messytable,seo2023vit} approaches utilize ground-truth bounding boxes and identity labels, and the self-supervised \cite{gan2021self} approach also uses ground-truth bounding boxes for training. For fair comparison and also to further evaluate our approach's capability of training with additional ground-truth data, we conduct an ablation study of (1) using only ground-truth bounding boxes and (2) using two types of annotations to train the proposed Self-MVA. Specifically, we replace the instance association in \cref{sec:syn} with the identity labels when provided. As shown in \cref{tab:gt}, our approach achieves remarkable performance, proving the effectiveness of the learning framework under different kinds of supervision.

\subsection{Applications}

\paragraph{\textbf{Automatic annotation}} Our self-supervised approach is capable of automatic multi-view identity annotation. We empirically evaluate such capability on the MVOR dataset: if we annotate the identity labels from scratch, it would take more than 6 hours; if we fine-tune the automatically generated identity labels, it would take less than an hour.

\paragraph{\textbf{Multi-object tracking}} To show that our approach is beneficial for multi-object tracking, we apply it on the Shelf \cite{belagiannis20143d} and Panoptic \cite{joo2015panoptic} datasets. As shown in \cref{fig:pose}, the accurate multi-view association provides strong cues for tracking when the person is completely occluded in one view.

\paragraph{\textbf{Multi-view multi-person pose estimation}} Our approach could potentially be used for multi-view multi-person pose estimation. \cref{fig:pose} shows an association example on the Panoptic dataset, which is a challenging benchmark for pose estimation. After the association, further 2d-3d pose lifting algorithm could be applied \cite{dong2019fast,xu2022multi} to obtain the 3d poses.
\section{Conclusion}

We present a self-supervised approach, Self-MVA, to address the challenging problem of multi-view person association in challenging scenes where appearance features are unreliable and where no annotations are available. Existing self-supervised approaches are limited to training a visual descriptor only, while we incorporate both appearance and geometric information for a more robust association. Using multi-view synchronization, we propose a self-supervised pretext task named cross-view image synchronization and design an encoder-decoder learning framework to encode fine-grained geometric features for each person, along with two types of self-supervised linear constraints to reduce the solution space. Experiments on three challenging datasets proves the effectiveness of our approach. However, our approach is not applicable to scenes with moving cameras, which presents an interesting direction for future research. 
\section{Acknowledgements}
This work was partially supported by French state funds managed by BPI France (project 5G-OR) and by the ANR under reference ANR-10-IAHU-02 (IHU Strasbourg). This work was also granted access to the servers/HPC resources managed by CAMMA, IHU Strasbourg, Unistra Mesocentre, and GENCI-IDRIS [Grant 2021-AD011011638R3].

{
    \small
    \bibliographystyle{ieeenat_fullname}
    \bibliography{main}
}
\clearpage
\setcounter{page}{1}
\maketitlesupplementary




\section{Additional experiments}
\label{sec:add_exp}

\subsection{Qualitative results}

\begin{figure*}[t]
\centering
\includegraphics[width=2.0\columnwidth]{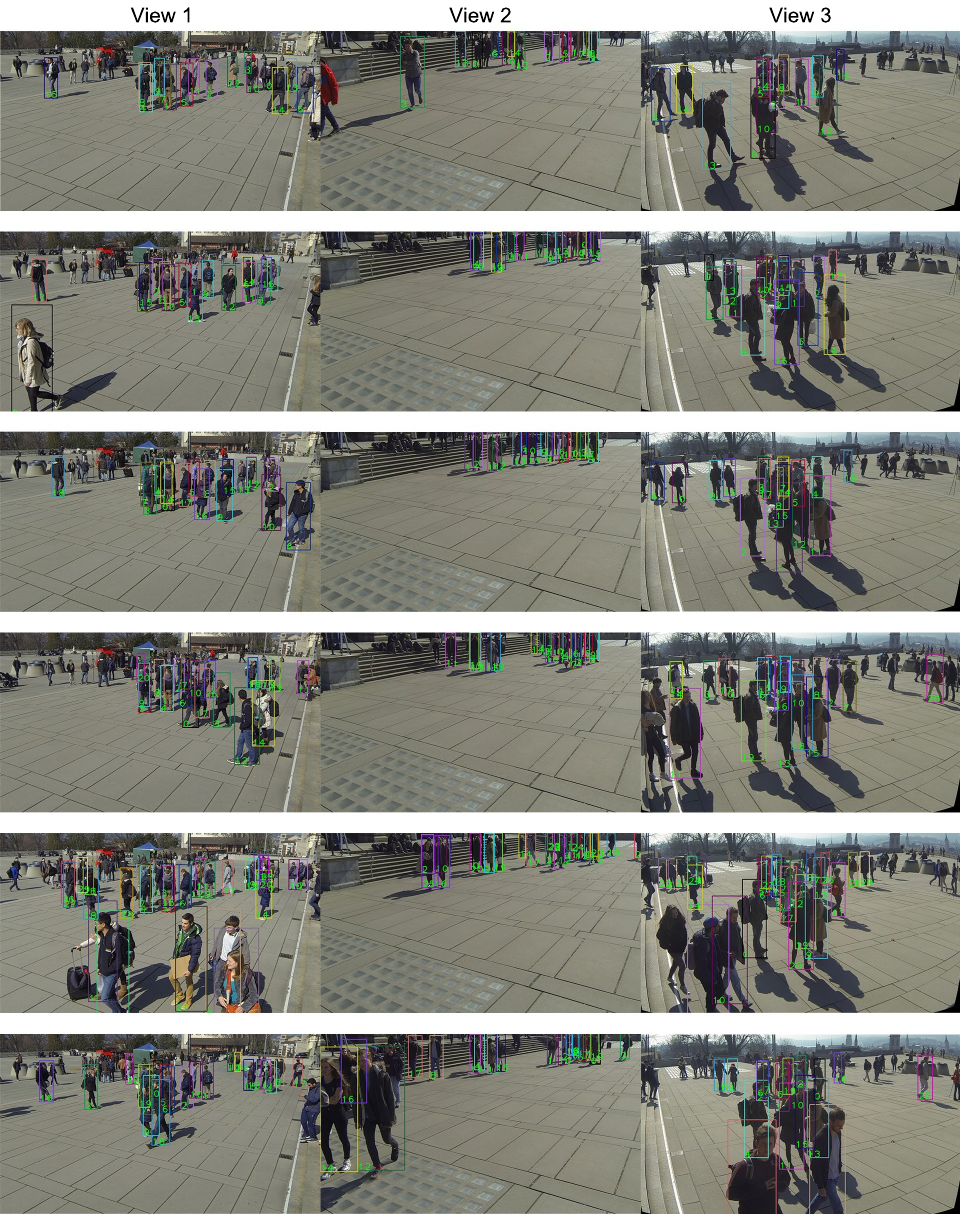} 
\caption{Qualitative results of our multi-view association approach on the WILDTRACK dataset (first 3 views).}
\label{fig:wildtrack}
\end{figure*}

\begin{figure*}[t]
\centering
\includegraphics[width=2.0\columnwidth]{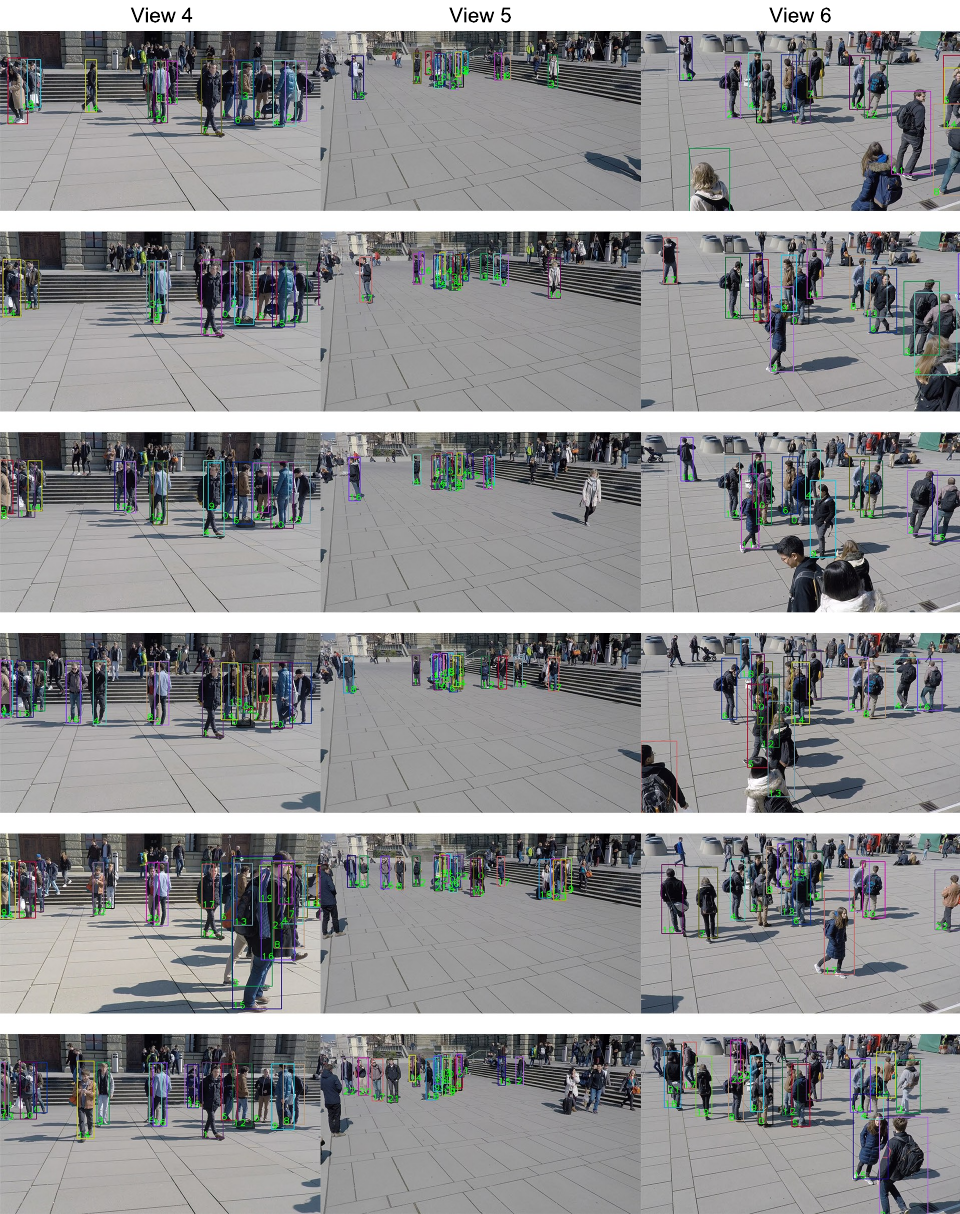} 
\caption{Qualitative results of our multi-view association approach on the WILDTRACK dataset (last 3 views).}
\label{fig:wildtrack2}
\end{figure*}

\begin{figure*}[t]
\centering
\includegraphics[width=2.0\columnwidth]{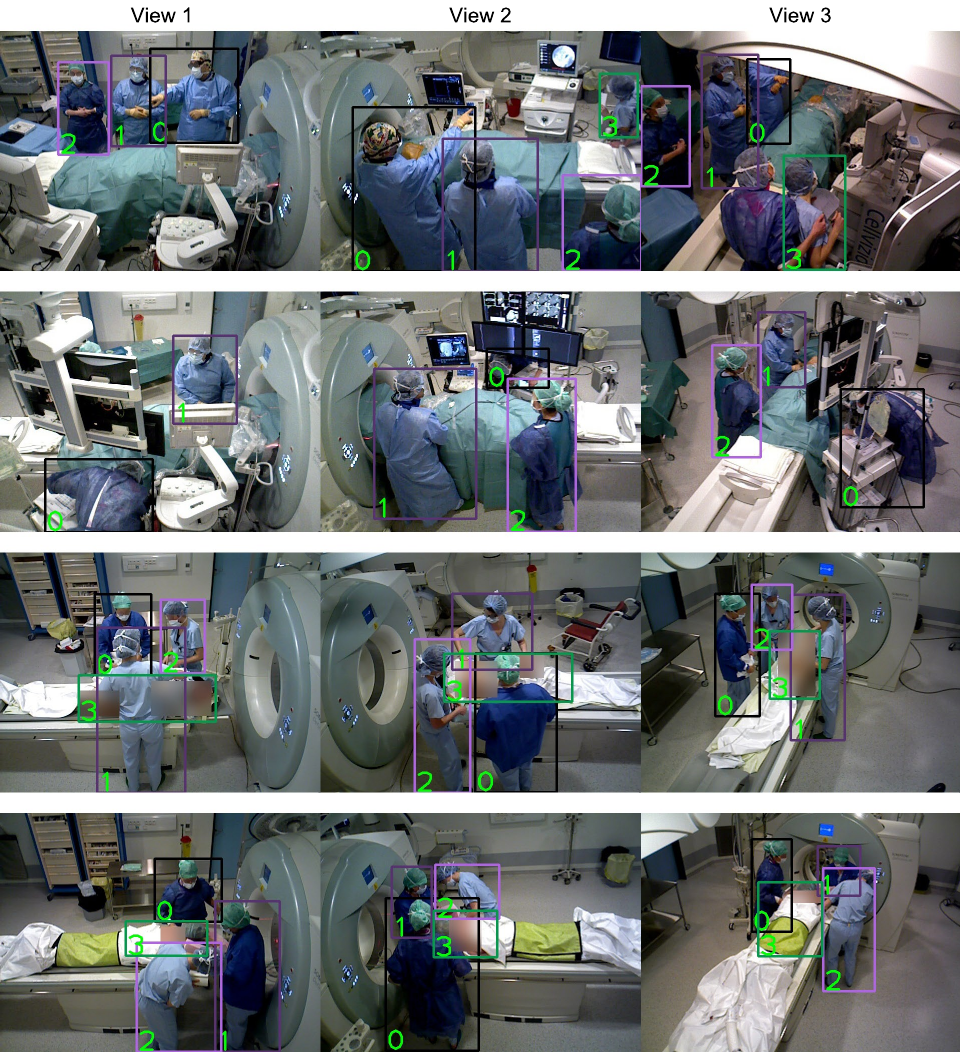} 
\caption{Qualitative results of our multi-view association approach on the MVOR dataset.}
\label{fig:mvor}
\end{figure*}

\begin{figure*}[t]
\centering
\includegraphics[width=2.0\columnwidth]{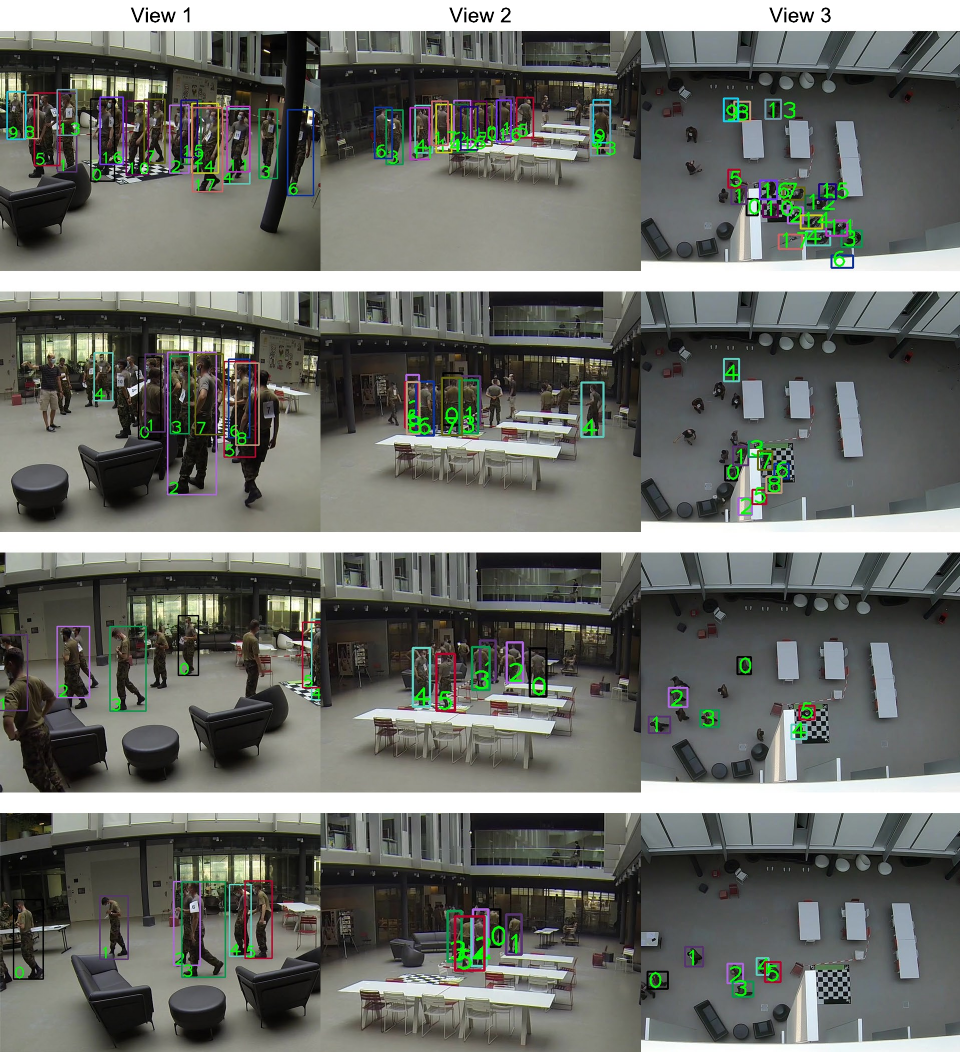} 
\caption{Qualitative results of our multi-view association approach on the SOLDIERS dataset (first 3 views).}
\label{fig:soldiers1}
\end{figure*}

\begin{figure*}[t]
\centering
\includegraphics[width=2.0\columnwidth]{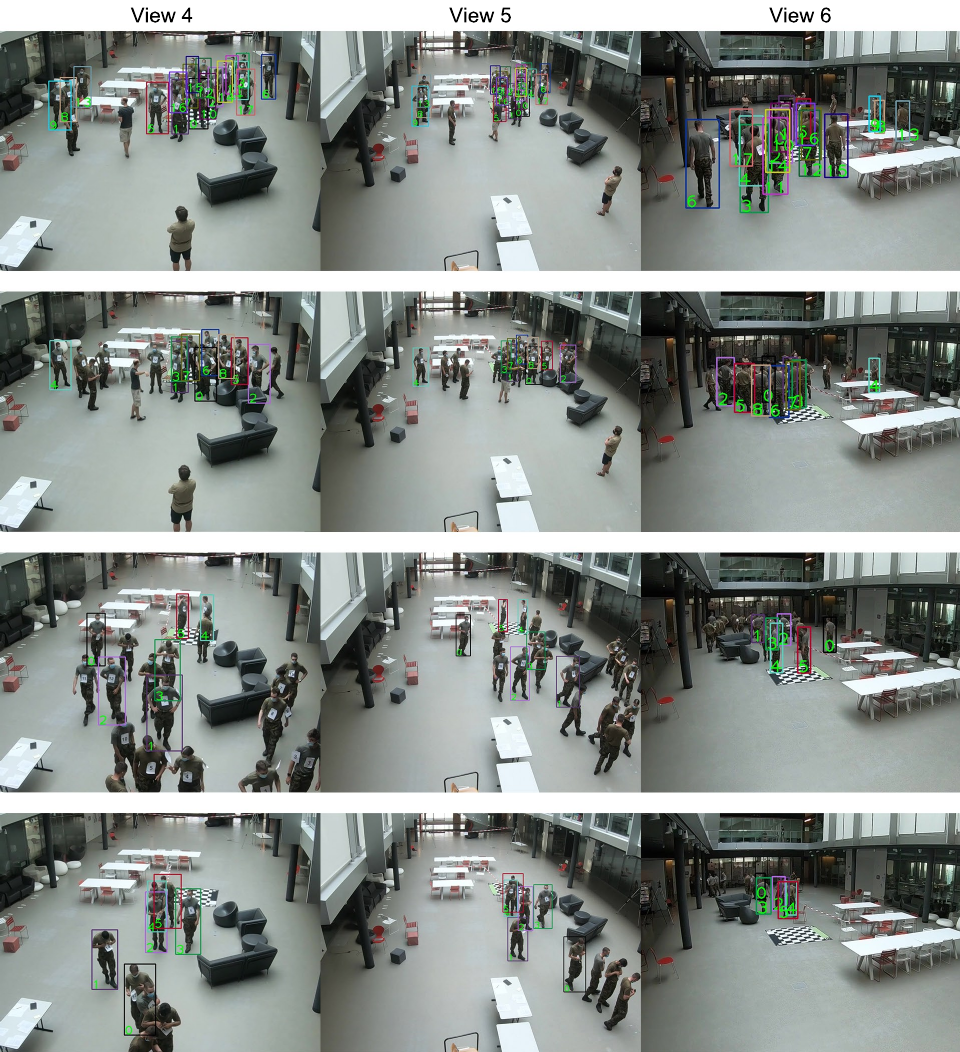} 
\caption{Qualitative results of our multi-view association approach on the SOLDIERS dataset (last 3 views).}
\label{fig:soldiers2}
\end{figure*}

We show more qualitative multi-view association results of our approach on the WILDTRACK~\cite{chavdarova2018wildtrack}, MVOR~\cite{srivastav2018mvor} and SOLDIERS~\cite{soldiers} datasets, as shown in \cref{fig:wildtrack,fig:wildtrack2,fig:mvor,fig:soldiers1,fig:soldiers2}. 

\subsection{Ablation study}

\paragraph{\textbf{Analysis of fusing Re-ID and geometric distances}} 
To investigate how the fusion of the Re-ID and geometric distances affects the performance, we conduct an ablation study of the $\alpha$ in \cref{eq:dis}. As shown in \cref{tab:alpha}, when $\alpha$ increases to 0.3 and 0.4, the performance starts to decrease on the WILDTRACK dataset. Although different datasets may have different optimal $\alpha$, the variance of the performance is not large, and thus it is easy to obtain decent performance by setting $\alpha$ hierarchically. 

\begin{table}[t]
\centering
\resizebox{1.0\columnwidth}{!}{
\begin{tabular}{c|cccccccccc}
    \toprule
    $\alpha$ & AP $\uparrow$ & FPR-95 $\downarrow$ & P $\uparrow$ & R $\uparrow$ & \textbf{ACC} $\uparrow$ & \textbf{IPAA-100} $\uparrow$ & \textbf{IPAA-90} $\uparrow$ & \textbf{IPAA-80} $\uparrow$ \\
    \midrule
    0 & 56.36 & 11.14 & 90.50 & 93.13 & 90.50 & 46.79 & 62.38 & 83.81 \\
    0.1 & 56.68 & 11.19 & 92.31 & \textbf{94.34} & 91.73 & \textbf{54.64} & 65.60 & 86.55 \\
    0.2 & 60.03 & \textbf{9.39} & \textbf{93.26} & 93.55 & \textbf{93.14} & 53.81 & \textbf{72.97} & \textbf{91.07} \\
    0.3 & \textbf{63.06} & 9.55 & 91.33 & 93.52 & 91.91 & 52.50 & 67.26 & 87.50 \\
    0.4 & 60.21 & 11.13 & 88.19 & 90.08 & 89.35 & 42.38 & 56.79 & 82.38 \\
    \bottomrule
\end{tabular}
}
\caption{Ablation study of the $\alpha$ in \cref{eq:dis}.}
\label{tab:alpha}
\end{table}

\paragraph{\textbf{Analysis of the self-supervised tasks}}  To further illustrate the roles of the self-supervised tasks, we display the ablation study on the MVOR and SOLDIERS datasets. As shown in \cref{tab:syn_loc_mvor_soldiers}, training with cross-view image synchronization task solely already achieves great performance, because these two datasets are relatively easier compared to the WILDTRACK dataset with a more crowded scene, leading to larger solution space for the association. Therefore, we need further linear constraints to reduce the solution space. \cref{tab:syn_loc_edge} shows that adding multi-view re-projection constraint significantly improves the training. 

\begin{table}[t]
\centering
\resizebox{1.0\columnwidth}{!}{
\begin{tabular}{c|c|c|cccccc}
    \toprule
    Data & Syn. & Pro. & AP $\uparrow$ & FPR-95 $\downarrow$ & P $\uparrow$ & R $\uparrow$ & ACC $\uparrow$ & IPAA-100 $\uparrow$ \\
    \midrule
    \parbox[b]{5mm}{\multirow{3}{*}{\rotatebox[origin=c]{90}{\makecell{MVOR}}}} & \checkmark & \xmark & 80.08 & 92.29 & 91.90 & \textbf{95.78} & 88.46 & \textbf{84.42}  \\
    & \xmark & \checkmark & 28.33 & 94.39 & 35.56 & 37.99 & 38.46 & 35.68  \\
    & \checkmark & \checkmark & \textbf{86.50} & \textbf{79.44} & \textbf{93.20} & 93.93 & \textbf{89.38} & 83.92 \\
    \midrule
    \parbox[b]{5mm}{\multirow{3}{*}{\rotatebox[origin=c]{90}{\makecell{SOLD.}}}} & \checkmark & \xmark & 69.66 & \textbf{32.14} & 92.17 & 92.17 & 92.17 & 77.82  \\
    & \xmark & \checkmark & 11.33 & 98.86 & 29.55 & 29.55 & 29.55 & 13.45  \\
    & \checkmark & \checkmark & \textbf{79.13} & 32.26 & \textbf{95.89} & \textbf{95.89} & \textbf{95.89} & \textbf{87.64} \\
    \bottomrule
\end{tabular}
}
\caption{Ablation study of the self-supervised learning tasks, including cross-view image synchronization (Syn.) and multi-view re-projection (Pro.). On the MVOR and SOLDIERS (SOLD.) datasets, applying cross-view image synchronization task already achieves great performance.}
\label{tab:syn_loc_mvor_soldiers}
\end{table}

\paragraph{\textbf{Confidence threshold}} Treating the cross-view association as a maximum bipartite matching problem may introduce false positive matches, and thus we propose to set confidence threshold for scores computed in \cref{eq:conf} to filter out the low-score matches. As shown in~\cref{tab:conf}, setting the threshold to 0.4 effectively improves precision and maintains a similar recall rate. 

\begin{table}[t]
\centering
\resizebox{0.9\columnwidth}{!}{
\begin{tabular}{c|cccccccc}
    \toprule
    Threshold & P $\uparrow$ & R $\uparrow$ & ACC $\uparrow$ & IPAA-100 $\uparrow$ & IPAA-90 $\uparrow$ & IPAA-80 $\uparrow$ \\
    \midrule
    0 & 89.55 & \textbf{94.54} & 88.23 & 46.43 & 56.19 & 77.50 \\
    0.2 & 90.35 & \textbf{94.54} & 89.39 & 49.88 & 59.52 & 80.60 \\
    0.4 & 92.31 & 94.34 & \textbf{91.73} & \textbf{54.64} & \textbf{65.60} & \textbf{86.55} \\
    0.6 & \textbf{96.00} & 79.74 & 85.90 & 17.50 & 49.88 & 76.07 \\
    \bottomrule
\end{tabular}
}
\caption{Ablation study of the confidence threshold on the WILDTRACK dataset.}
\label{tab:conf}
\end{table}

\subsection{Failure cases}

\begin{figure}[t]
  \centering
   \includegraphics[width=1.0\linewidth]{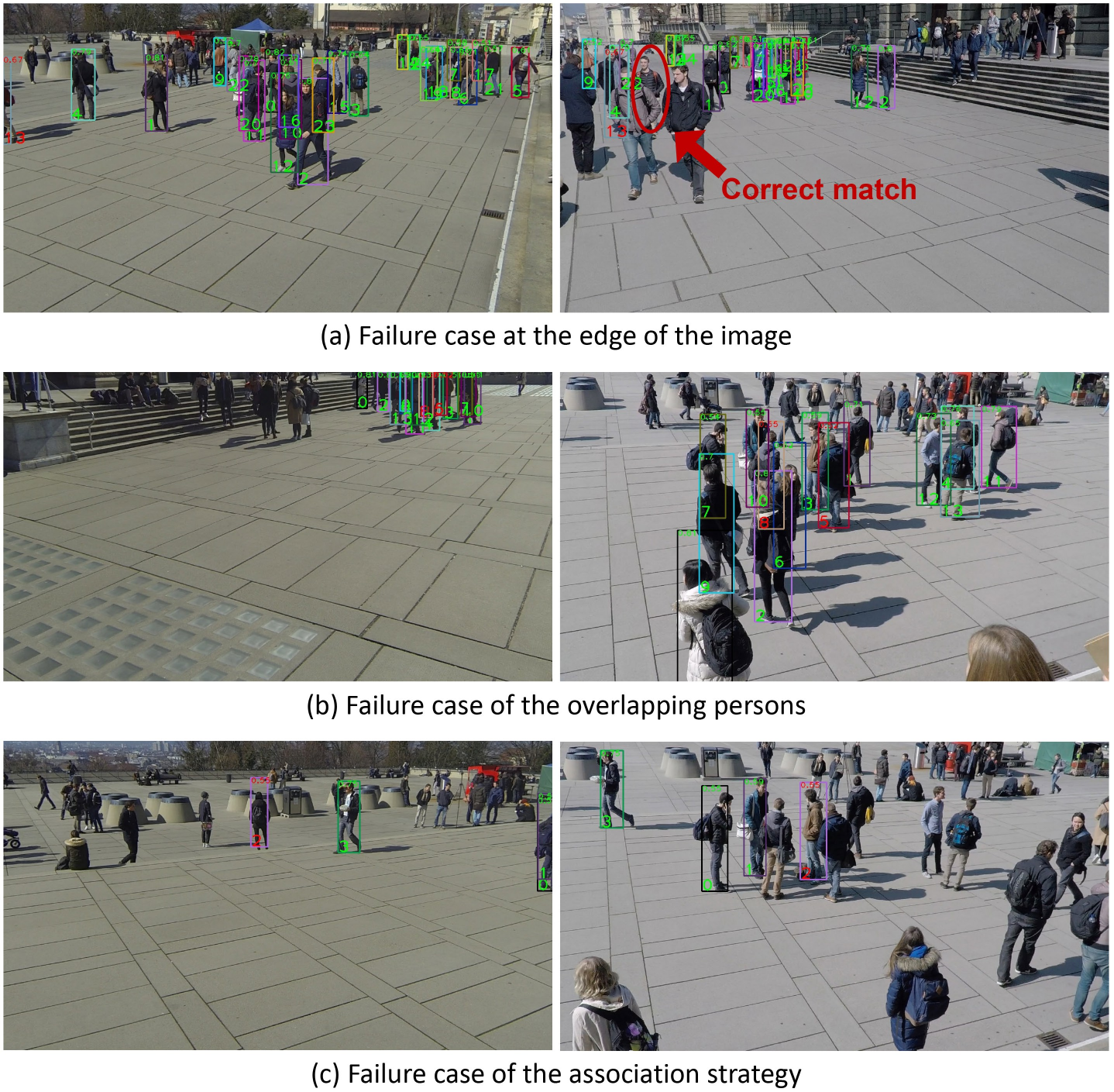}
   \caption{Examples of the failure cases: (a) the person at the edge of the image is incorrectly associated with the person standing next to him; (b) the two overlapping persons are associated with each other; (c) the two persons that only appear in one view are incorrectly associated. }
   \label{fig:failure}
\end{figure}

\cref{fig:failure} shows three types of failure cases from our approach on the WILDTRACK dataset. In \cref{fig:failure}a, the person at the edge of the image is severely occluded. Although Self-MVA manages to find its approximate position in the other view, it incorrectly associates him with the person standing next to him. In \cref{fig:failure}b, the two overlapping persons with obscure spatial relationships are incorrectly associated with each other. In \cref{fig:failure}c, the two persons that only appear in one view are incorrectly associated due to the logic of the Hungarian matching. 

\section{Limitations and future work}

Although our work achieves great performance on different challenging benchmarks in a self-supervised way, there are some limitations. 

First, we only solve the self-supervised multi-view person association with stationary cameras. For the other scenes where the cameras are continuously moving, our approach does not perform well, because the camera embeddings and the decoder for each view are fixed. To conduct self-supervised multi-view association with the moving cameras, there are two possible directions: (1) to model the continuous camera poses by predicting the relative camera poses in adjacent frames; (2) to solve the association in a one-shot setting by strong 3d geometric reasoning. 

Second, we use a manually defined threshold to filter out the false matches for each dataset during the inference. Specifically, we select the threshold from $\left\{ 0.0, 0.1, 0.2, ..., 0.9\right\}$ based on the results on the validation set. For datasets without labels, we need to adjust the threshold value by manually observing the association results for better performance. Even if we have set a good threshold value, the false positive matches still exist as shown in \cref{fig:failure}c. Therefore, how to automatically and accurately remove the false positive matches during inference remains unexplored in our work. To effectively detect these false positive matches, explicitly calculating the epipolar constraint between the two views would be a possible solution.


\end{document}